\documentclass[runningheads]{llncs}

\let\llncssubparagraph\subparagraph
\let\subparagraph\paragraph
\usepackage[compact]{titlesec}
\let\subparagraph\llncssubparagraph

\usepackage{eccv}

\usepackage{eccvabbrv}

\usepackage{graphicx}
\usepackage{booktabs}
\usepackage{comment}
\usepackage{soul}
\usepackage{titlesec}
\usepackage{booktabs}
\usepackage{graphicx}
\usepackage{multirow} 
\usepackage{makecell}
\usepackage{tikz}
\usepackage{arydshln}
\usepackage{float}
\usepackage{enumitem}

\usepackage[accsupp]{axessibility}  
\usepackage{caption}

\usepackage{hyperref}

\usepackage{orcidlink}

\begin{document}

\title{LAR-IQA: A Lightweight, Accurate, and Robust No-Reference Image Quality Assessment Model} 

\titlerunning{LAR-IQA}

\author{Nasim Jamshidi Avanaki\inst{}\orcidlink{0009-0006-9127-6249} \and
Abhijay Ghildyal\inst{1}\orcidlink{0000-0003-1940-9626} \and
Nabajeet Barman\inst{2}\orcidlink{0000-0003-2587-7370} \and
Saman Zadtootaghaj\inst{2}\orcidlink{0000-0002-6028-8507}}

\authorrunning{Jamshidi Avanaki et al.}

\institute{Portland State University, OR, USA \and Sony Interactive Entertainment, UK/Germany\\
\email{n.jamshidi.avanaki@gmail.com},
\email{abhijay@pdx.edu}\\
\email{\{Nabajeet.Barman,Saman.Zadtootaghaj\}@sony.com}}

\maketitle

\begin{abstract}

Recent advancements in the field of No-Reference Image Quality Assessment (NR-IQA) using deep learning techniques demonstrate high performance across multiple open-source datasets. However, such models are typically very large and complex making them not so suitable for real-world deployment, especially on resource- and battery-constrained mobile devices. To address this limitation, we propose a compact, lightweight NR-IQA model that achieves state-of-the-art (SOTA) performance on ECCV AIM UHD-IQA challenge validation and test datasets while being also nearly 5.7 times faster than the fastest SOTA model. Our model features a dual-branch architecture, with each branch separately trained on synthetically and authentically distorted images which enhances the model's generalizability across different distortion types. To improve robustness under diverse real-world visual conditions, we additionally incorporate multiple color spaces during the training process. We also demonstrate the higher accuracy of recently proposed Kolmogorov-Arnold Networks (KANs) for final quality regression as compared to the conventional Multi-Layer Perceptrons (MLPs). Our evaluation considering various open-source datasets highlights the practical, high-accuracy, and robust performance of our proposed lightweight model. Code: \url{https://github.com/nasimjamshidi/LAR-IQA}.


\end{abstract}

\section{Introduction}
\label{sec:intro}

The Image Quality Assessment (IQA) task of measuring the quality of images as perceived by humans remains one of the most interesting and challenging fields in computer vision. No-Reference IQA (NR-IQA), also known as Blind IQA (BIQA), focuses on estimating the quality of degraded images when there is no high-quality reference image available for comparison. This task is particularly challenging given the wide range of possible distortions (compression, blur, noise, etc.) that might be present in an image. NR-IQA plays a critical role across multiple industries and applications, such as photography, surveillance, healthcare, automotive, social media, and user-generated content platforms. Millions of user-generated content (UGC) images are uploaded daily and shared across numerous social media platforms such as Instagram, X, and Flickr. In the wild, user-captured images can suffer from distortions such as blurriness, noise (from the camera sensor), color distortions, compression artifacts (blockiness), or a combination of these issues. Automatically detecting low-quality or inappropriate images and guiding the necessary pre- and post-processing steps (quality enhancement, compression factor, deblurring, etc.) is critical for enhancing user experience and the success of such companies.

One of the major challenges in NR-IQA is developing smaller, faster and more efficient methods suitable for real-time quality assessment. Traditional NR-IQA models are generally fast and less complex but often lack accuracy~\cite{zhang2015feature, mittal2012making, mittal2012no, moorthy2011blind, xue2014blind, ye2012unsupervised, xu2016blind, saad2012blind, liu2014no, moorthy2010two}. On the other hand, traditional DNN-based models, while more accurate, typically have higher complexity and are computationally intensive~\cite{agnolucci2024arniqa, bodner2024convolutional, yang2022maniqa, su2020blindly, golestaneh2022no, yun2023uniqa, wang2023exploring}. Recent IQA models based on large multi-modal models utilize sophisticated architectures such as transformers for both vision and text encoders~\cite{zhu2024adaptive,qalign} resulting in large model size and complexity, making them unsuitable for most real-time evaluation tasks. 

Some more recent NR-IQA methods use Transformers as their backbone network~\cite{yang2022maniqa, golestaneh2022no, yun2023uniqa} since Transformers have been shown to provide better features for NR-IQA, resulting in more accurate and robust results. However, Transformer-based models consists of a large number of parameters and require significant computational resources, both for training and inference, thus restricting their applicability for deployment on low-power devices. Recently, Vision Language Models (VLMs), particularly CLIP~\cite{radford2021learning}, have achieved significant success in NR-IQA. When fine-tuned for NR-IQA tasks~\cite{wang2023exploring, agnolucci2024quality, he2024cover}, CLIP demonstrates good accuracy, robustness, and generalization compared to existing methods, effectively capturing a wide range of diverse distortions in large datasets. However, we do not utilize VLM-based pre-trained networks and instead focus on developing an accurate and robust model that is more computationally efficient, specifically in terms of the number of Multiply-Accumulate (MAC) operations required for a forward pass with an input image size of 3840 × 2160 pixels. 

\subsection{Contributions}

This paper makes several contributions for low complexity, generalizablity and robustness which is discussed next. 

\subsubsection{Lower Complexity.}
Due to the ability of DNN-based methods to capture intricate patterns and non-linear distortions for higher accuracy and robustness, we use the lightweight MobileNetV3~\cite{howard2019searching} as the baseline network. Compared to other DNN-based NR-IQA methods~\cite{agnolucci2024arniqa, su2020blindly, wang2023exploring, agnolucci2024quality}, our network has fewer parameters, resulting in lower complexity.

\subsubsection{Generalizability.}
One of the main challenges in training a generalizable NR-IQA model is that synthetic datasets lack the diverse distortions found in real-world scenarios. Several methods have attempted to address this gap using pre-training, self-supervision, and novel loss functions~\cite{agnolucci2024arniqa,golestaneh2022no,wang2023exploring, agnolucci2024quality}. Self-supervised learning with contrastive loss or pre-training on unlabeled data~\cite{agnolucci2024arniqa, madhusudana2022image} is effective for handling both synthetic and authentic types of degradation. However, we address the problem of lack of comprehensive datasets by training two separate models — one on dataset with synthetic distortions and  another on authentic distortion datasets. Both individually trained models are then combined to create a more generalized model. 

Such multiple branch architectures have been proposed previously. For example, authors in~\cite{he2024cover} used three branches: semantic, aesthetic, and technical, the prediction scores from which are then fused together to obtain the final quality score. In contrast, ours is the first work to propose a dual-branch architecture to tackle two different types of distortions, synthetic and authentic. Given the need for real-time applications, we focus on making our NR-IQA model both lightweight and fast by focusing exclusively on MobileNet~\cite{howard2019searching, zhao2022new, vasu2024mobileclip, luo2020comparison} as the backbone image encoder for our model. 

\subsubsection{Robustness and Accuracy.}
To enhance our lightweight model’s robustness and accuracy, we incorporate several strategies: separate dual branch training (synthetic and authentic distortions), using  KAN~\cite{liu2024kankolmogorovarnoldnetworks} as the regression head, and using a color space loss function.

In summary, this paper introduces a novel approach and model for low-complexity NR-IQA, leveraging both authentic and synthetic distortions for robust and accurate evaluations. Our key contributions are:
\begin{itemize}[noitemsep, topsep=0pt]
    \item We propose combining authentic and synthetic IQA branches, each trained on different datasets, to enhance the model's robustness and generalizability across various distortions.
    \item We compare different backbone architectures tailored for the synthetic and authentic branches, facilitating optimal design choices.
    \item We conduct an ablation study comparing KANs and MLPs for the quality regression module. To the best of our knowledge, this is the first study to investigate and compare the efficacy of KANs against MLPs in this context.
    \item We propose a novel loss function addressing variations in different color spaces to improve the robustness of the IQA models.
\end{itemize}

\noindent Our approach results in a lightweight model that surpasses state-of-the-art performance~\cite{hosu2024uhd} on the validation and test datasets of the ECCV AIM UHD-IQA challenge\footnote{\url{https://codalab.lisn.upsaclay.fr/competitions/19335}} while maintaining efficiency for practical applications. 

The rest of the paper is organized as follows: In Section~\ref{sec:Rel}, we discuss prior work related to No-Reference Image Quality Assessment and the integration of synthetic and authentic data. In Section~\ref{sec:PM}, we provide a detailed description of our model architecture and the comparative analysis of different backbones, heads, and loss functions. Section~\ref{sec:Exp} presents the results of our ablation study and the evaluation of our proposed model versus SOTA. Finally, Section~\ref{sec:Conc} concludes this paper by highlighting the implications of our findings.   

\section{Related Work} \label{sec:Rel}

\subsection{No-Reference Image Quality Assessment}

Over the years, many NR-IQA models have been proposed, from initial traditional approaches based on Natural Scene Statistics (NSS) and hand-crafted features to deep-learning based approaches to more recently, models based on Vision-Language Models (VLMs). Traditional NR-IQA approaches relied on the detection and measurement of specific distortions such as blockiness, blur, banding, and noise, among others. Such methods leveraged hand-crafted natural scene statistics (NSS) features to assess image quality~\cite{ssim, zhang2015feature, mittal2012making, mittal2012no, moorthy2011blind, xue2014blind, ye2012unsupervised, xu2016blind, saad2012blind, liu2014no, moorthy2010two}. However, these distortion-specific models based on hand-crafted features do not generalize well to images with multiple types of distortions. 

Recently, driven by the success of deep neural networks (DNNs) in computer vision, several NR-IQA methods have been proposed that generalize better without the need for explicitly designing features. DNN-based approaches achieve high accuracy across various datasets by learning complex patterns and accounting for multiple distortions in images~\cite{agnolucci2024arniqa,bodner2024convolutional,yun2023uniqa,yang2022maniqa}. However, due to the limited size of available datasets, they often tend to overfit on training data and cannot generalize well to newer distortions (not present in the training data), which are evident when performing cross-dataset evaluation as discussed in~\cite{golestaneh2022no}. 

More recently, Vision-Language Models (VLMs) have found success in the field of IQA, combining both visual and textual information to assess the image quality~\cite{wang2023exploring}. Since VLMs have been trained on very large-scale datasets that combine vision and language modalities, they have a more nuanced understanding of the image content using its context. A combination of semantic information and textual descriptions/captions allows them to better understand the quality, often resembling human ratings. Similarly, large multi-modality models (MLLM) have been developed for the IQA task \cite{zhu2024adaptive, qalign} and are more accurate than traditional DNN-based methods. However, their complex training setups and large scale make VLM and MLLM-based models computationally expensive.

\subsection{Generalization Across Diverse Datasets} \label{sec:rel_gen_dataset}

Training large Vision Transformer (ViT) or CNN backbone networks requires extensive datasets for pre-training to ensure generalization to new datasets. While some approaches utilize contrastive learning for pre-training~\cite{agnolucci2024arniqa, madhusudana2022image}, others attempt to merge multiple image datasets with available subjective scores~\cite{mittag2021bias}. However, merging datasets is challenging due to inherent subjective biases within each dataset. As described in~\cite{mittag2021bias}, these biases include rating noise, subjective test order effects, varying distributions of quality scores across datasets, and long-term dependencies in subjective experiments. Additional biases, especially when crowdsourcing is used, arise from less controlled environments, such as differences in monitor distances, display sizes, and settings (e.g., gamma and luminance variations). These biases complicate the straightforward merging of subjective scores from different tests.

To address these challenges, various alternative methods have been proposed, such as integrating subjective biases into the loss function~\cite{mittag2021bias}, and using ranking loss instead of Mean Squared Error (MSE) loss, based on the assumption that subjective biases do not affect the ranking of MOS within a dataset~\cite{golestaneh2022no}. We on the other hand, train each branch of our model using multiple authentic and synthetic datasets by employing a multi-task training approach. For each branch, each dataset is considered as a separate task with a unique head. This allows each branch to effectively learn the subjective biases present in respective individual datasets leading to a better generalization of the image encoder. In the final model, we remove these individual heads, and instead use a single KAN head for each of the two branches as a down-sampler, and then fuse the resulting embeddings with a larger KAN head.

\subsection{Kolmogorov-Arnold Networks (KAN)}

While typically fully-connected MLPs have been used as regression heads~\cite{he2024cover, cheon2021perceptual}, recently, KANs ~\cite{liu2024kankolmogorovarnoldnetworks} have been introduced as an alternative to MLPs. Unlike MLPs, which multiply the input by weights, KANs process the input through learnable B-spline functions~\cite{bodner2024convolutional}. In the intermediate layers, while MLPs use a fixed activation function as $\sigma$ in $\sigma (w . x + b)$, KANs employ learnable activation functions based on parameterized B-spline functions, formulated as $\phi(x) = silu(x) + \sum_i c_i B_i(x)$, where $B$ is the basis function and $c$ is the trainable control point parameter. In this combination, the non-trainable component $silu(x)$ is added to the trainable spline component, transforming $\phi(x)$ into a residual activation function. Therefore, KAN consists of spline-based univariate functions along the network edges, designed as learnable activation functions. These features enhance both the accuracy and interpretability of the network. In our work, replacing the MLP head with KAN also enhances the accuracy of our model, as presented later in Section~\ref{sec:mlpVsKan}.
    
\section{Proposed Method} \label{sec:PM}

\begin{figure}[t]
    \centering
    \includegraphics[width=\textwidth]{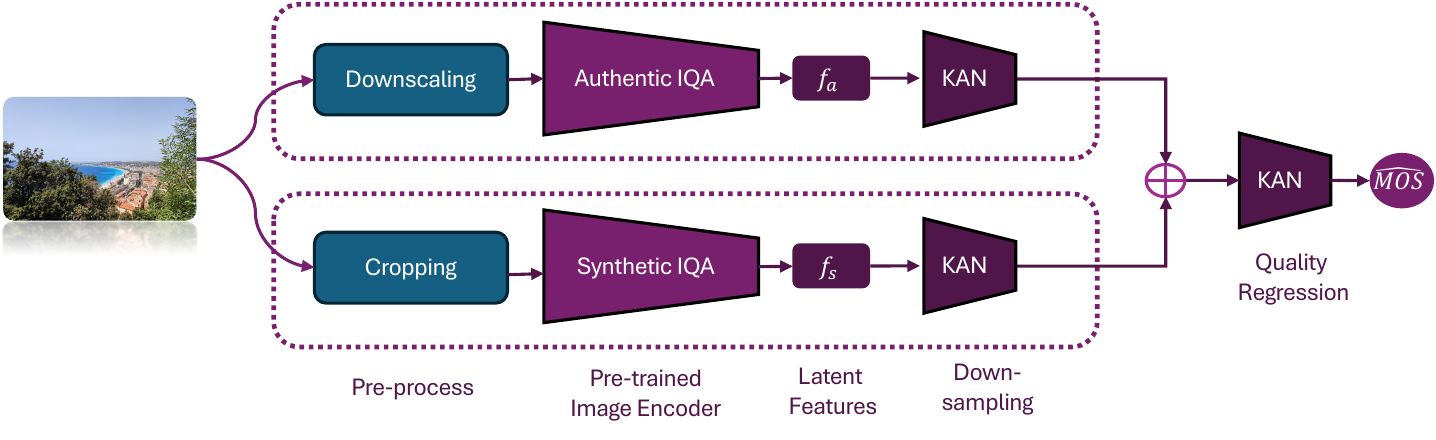}
    \caption{Proposed model architecture. The image quality is evaluated using two branches: Authentic and Synthetic. In both branches, MobileNetV3~\cite{howard2019searching} serves as the lightweight image encoder. The features extracted from these branches are concatenated and then used as input to KAN, the quality regression module, which outputs the final predicted image quality score.}
    \label{fig:mainFigure}
\end{figure}

In this paper, we aim to develop a lightweight model, focusing on both Multiply-Accumulate Operations (MACs) and the number of parameters. To achieve this, we utilize mobile architectures~\cite{luo2020comparison} in our image encoder. We propose a dual-branch architecture, comprising Authentic and Synthetic branches, as shown in Figure~\ref{fig:mainFigure}.
Each branch includes a pre-processing module and a mobile-based image encoder. We use a KAN quality regression module to merge features from both branches and predict the final quality score. Initially, the two branches are treated as separate models and trained on authentic and synthetic datasets, respectively. Each branch employs a KAN regression head to independently predict image quality. After training the branches on their respective datasets, the image encoders from both the Authentic and Synthetic branches are integrated into the dual-branch model.

    \subsection{Datasets and Evaluation Metrics}
    
    In this work, we use seven publicly available IQA datasets: two with synthetic distortions (KADID-10K~\cite{lin2019kadid}, TID2013~\cite{ponomarenko2015image}), four with authentic distortions (KonIQ-10k~\cite{hosu2020koniq}, SPAQ~\cite{fang2020perceptual}, BID~\cite{ciancio2010no}, and UHD-IQA~\cite{hosu2024uhd}), the PIPAL~\cite{jinjin2020pipal} with both synthetic and algorithmic distortions. A summary of the datasets used in our experiments is provided in Table~\ref{tab:datasets}.

    To evaluate the model's performance, we use three common criteria: Spearman Rank Order Correlation Coefficient (SRCC) for prediction monotonicity, Pearson Linear Correlation Coefficient (PLCC) and Kendall Rank Correlation Coefficient (KRCC) for rank consistency.

\begin{table}[tb]
\centering
\caption{Summary of IQA datasets used in this work.}
\label{tab:datasets}
\scriptsize
\resizebox{\textwidth}{!}{%
\begin{tabular}{@{}lccccc@{}}
\specialrule{0.75pt}{0pt}{0pt} 
Database & 
  \begin{tabular}[c]{@{}c@{}}\# of Source \\ Images\end{tabular} & 
  \begin{tabular}[c]{@{}c@{}}Dist. \\ Type\end{tabular} & 
  \begin{tabular}[c]{@{}c@{}}\# of Dist. \\ Images\end{tabular} & 
  \begin{tabular}[c]{@{}c@{}}\# of Dist. \\ Types\end{tabular} & 
  \begin{tabular}[c]{@{}c@{}}Image \\ Resolution\end{tabular} \\ \hline
KONIQ-10K~\cite{hosu2020koniq} & 10,073 & Authentic & 10,073 & -- & 1024×768px \\
SPAQ~\cite{fang2020perceptual} & 11,125 & Authentic & 11,125 & -- & Variable \\
BID~\cite{ciancio2010no} & -- & Authentic & 585 & -- & 1280×960px to 2272×1704px \\
UHD-IQA~\cite{hosu2024uhd} & 6,073 & Authentic & -- & -- & mostly 3840×2160px \\
KADID-10k~\cite{lin2019kadid} & 81 & Synthetic & 10,125 & 25 & 512×384px \\
TID2013~\cite{ponomarenko2015image} & 25 & Synthetic & 3,000 & 24 & 512×384px \\
PIPAL~\cite{jinjin2020pipal} & 250 & Synthetic+Algorithmic & 29,000 & 40 & Variable \\
\bottomrule
\end{tabular}%
}
\normalsize
\end{table}
\subsection{Data Pre-processing}
For each branch of our model, we applied distinct pre-processing methods during the training phase. In the Authentic branch, images were downscaled to 224x224 pixels for pretraining and 384x384 for UHD-IQA challenge, while in the Synthetic branch, images were kept at their original size, and only the cropping operation was applied. The rationale for downscaling in the Authentic branch is that the quality of Authentic datasets is intrinsically linked to their content (e.g., captured objects, image composition), and changes in image size do not impact perceived quality~\cite{sun2023blind,imgDistortion}. However, in the Synthetic branch, downscaling can cause the loss of high-frequency details, which negatively impacts the predicted quality for images with synthetic distortions. Therefore, we only downscale images in the Authentic branch. For the Synthetic branch, we trained our model using cropped images of 224 sizes.
    
\subsection{Quality Regression Module}

For quality regression head we explore both MLPs and KANs. For KAN, we use the implementation provided by~\cite{Efficient-KAN}. To ensure a fair comparison between the MLPs and KANs, we conducted two sets of comparisons. First, both models were evaluated using an identical architecture, consisting of two fully connected (FC) layers: the first layer reduced 1000 features to 128, followed by a second layer that reduced 128 features to 1. Given that KAN uses a higher number of parameters and FLOPs, we also adjusted the MLP model's complexity to roughly match KAN's FLOPs by adding an additional layer with 1125 neurons before the 128-neuron layer, using ReLU activation. The evaluations are conducted under two scenarios: within-dataset evaluation and cross-dataset evaluation, with further details discussed in Section~\ref{sec:mlpVsKan}.

\subsection{Image Encoder Module}
For image encoder module, we explore two lightweight CNN-based architectures and two ViT architectures. As demonstrated in~\cite{luo2020comparison}, MobileNetV2~\cite{sandler2018mobilenetv2} and MobileNetV3~\cite{howard2019searching} are excellent choices for mobile CNN-based networks due to their reduced number of parameters and low computational complexity. Additionally, we consider the recent mobile ViT model, MobileCLIP-S2~\cite{vasu2024mobileclip}, and MobileViT-S \cite{mehta2021mobilevit} known for its lightweight characteristics. It is important to note that although MobileCLIP-S2 is a VLM, in Section~\ref{sec:backbones}, we only utilize its vision encoder architecture, and our model does not follow a VLM training approach.

\subsection{Loss Functions}

\subsubsection{MSE and PLCC ($\mathcal{L}_{acc}$).}

Distance-based losses, such as MAE and MSE, minimize the absolute difference between predicted scores and ground truth labels, ensuring accuracy in score prediction. In contrast, correlation-based losses like PLCC preserve the relative correlation of image quality, aligning better with human perception~\cite{chen2024gmc}. In this work, we enhance model accuracy by incorporating both distance-based (MSE) and correlation-based (PLCC) objectives as a combined loss function in all our experiments. 

\begin{equation}
\mathcal{L}_{acc} = \alpha \cdot MSE + \beta \cdot PLCC
\end{equation}

\subsubsection{Color Space Robustness ($\mathcal{L}_{rob}$).}
As shown in traditional NSS-based models ~\cite{bampis2017speed, ghadiyaram2017perceptual, madhusudana2021st, tu2021rapique}, analyzing images in different color spaces improves the accuracy of visual quality assessments by providing deeper insights into perceptual attributes that affect image quality. The multi-color space approach captures diverse aspects of image quality that may not be covered by a single color space. Therefore, we propose a \textit{color space loss} that evaluates image quality in the RGB, YUV, and LAB color spaces, resulting in more accurate and robust assessments. Our \textit{color space loss} ($\mathcal{L}_{rob}$) is defined as
\begin{equation}
    \widehat{MOS}_{\text{color-space}} = F_\phi (\mathbf{I}_{\text{color-space}})
\end{equation}
\begin{equation}    
    \sum_\text{color-space} \frac{1}{N} (\widehat{MOS}_{\text{color-space}} - MOS)^2    
\end{equation}
\noindent where \(\text{color-space} \in \{\text{RGB}, \text{YUV},  \text{LAB}\}\), and \(\mathbf{I}_{\text{color-space}}\) represents the input image in the specified color space. Using our trained model $F$ with parameters \(\phi\), we predict the MOS for \(\mathbf{I}_{\text{color-space}}\), denoted by $\widehat{MOS}_{\text{color-space}}$. To compute the loss we calculate the mean squared error with the ground-truth MOS. By minimizing this \textit{color space loss}, the model learns to align its output feature across various color-space representations. This approach enhances the model's ability to predict image quality consistently, regardless of the color space used. It improves the model's sensitivity to different distortions and ensures that assessments closely match human visual perception (see Section~\ref{sec:loss}).

\section{Evaluation Methodology and Results} \label{sec:Exp}

In Section~\ref{sec:implementation}, we first discuss the implementation details of our model. We present the initial results of the Synthetic and Authentic models in Section~\ref{sec:mlpVsKan} and explore replacing the MLP regression head with KAN. In Section~\ref{sec:backbones}, we experiment with different backbone networks as the image encoder for our Synthetic and Authentic models while keeping the regression head fixed as KAN. To enhance robustness, we investigate various loss functions in Section~\ref{sec:loss}. Finally, we combine insights from these experiments and discuss the setup and integration of the Synthetic and Authentic branches in Section~\ref{sec:syn_auth}. In Section~\ref{sec:sota_comparison}, we compare the performance of our final model against state-of-the-art methods based on accuracy metrics (SRCC, PLCC, KRCC, RMSE, and MAE) and computational complexity (MACs).

\subsection{Implementation Details} \label{sec:implementation}

The model training and testing is done on Pytorch on a single Nvidia A100 GPU. The models are trained for 100 epochs using the AdamW optimizer, starting with a learning rate of \(5\times10^{-5}\) and a weight decay of \(1\times10^{-4}\). We use a custom scheduler, that includes a linear warmup phase followed by a cosine annealing schedule. The input image size varies depending on the branch: original size for the synthetic branch and resized 224x224 pixels for the authentic branch. For the UHD-IQA dataset, due to the large image resolutions, we resized images to 384x384 for the authentic branch and applied 1280x1280 center crops for the synthetic branch.
The image encoder module in our final proposed model utilizes MobileNetV3, which is consistent across both branches. For the quality regression modules, we incorporate the KAN model to reduce the feature dimension and predict image quality, respectively. \\

\subsection{Quality Regression Module: MLP versus KAN} \label{sec:mlpVsKan}

    \begin{table}[t]
      \setlength{\tabcolsep}{4pt}
      \caption{Within-dataset evaluation results for different regression heads, MLP and KAN. The synthetic model is trained on  KADID-10K and PIPAL datasets and evaluated using 10-fold cross-validation. Results indicate better performance of KAN compared to MLP on both datasets.}
      \label{tab:single-head}
      \centering
      \begin{tabular}{@{}lcccccc@{}}
        \toprule
        Dataset & \multicolumn{3}{c}{KADID-10K} & \multicolumn{3}{c}{PIPAL} \\
        \cmidrule(r){2-4} \cmidrule(r){5-7}
        Head & PLCC & SRCC & KRCC & PLCC & SRCC & KRCC \\
        \midrule
        MLP & 0.961 & \textbf{0.946} & 0.825 & 0.868 & 0.845 & 0.680 \\
        KAN & \textbf{0.965} & 0.941 & \textbf{0.857} & \textbf{0.887} & \textbf{0.860} & \textbf{0.692} \\
        \bottomrule
      \end{tabular}
    \end{table}

\begin{table}[tb]
  \setlength{\tabcolsep}{4pt} 
  \caption{Cross-dataset evaluation results for MLP and KAN as regression heads. Synthetic model is trained on the KADID-10K dataset and tested on the TID2013 dataset. Similarly,the Authentic model is trained on KONIQ-10K dataset and tested on BID dataset. Results indicate significantly better performance of KAN compared to MLP.}
  \vspace{-0.1in}
  \label{tab:cross-head}
  \centering
  \begin{tabular}{@{}lcccccc@{}}
    \toprule
    \multirow{2}{5em}{Model} & \multicolumn{3}{c}{Head} &\multirow{2}{4em}{PLCC} & \multirow{2}{4em}{SRCC} & \multirow{2}{4em}{KRCC} \\
    \cmidrule(r){2-4}
    &MLP &KAN &FLOPs & & & \\
    \midrule
    & & & &\multicolumn{3}{c}{KADID-10K / TID2013} \\
    \cmidrule(r){5-7}
    \multirow{3}{5em}{Synthetic} &\checkmark & & 0.26M & 0.64 & 0.59 & 0.45 \\
    &\checkmark & & 2.60M & 0.67 & 0.63 & 0.48 \\
    & &\checkmark & 2.31M & \textbf{0.71} & \textbf{0.66} & \textbf{0.50} \\
    \midrule
    & & & &\multicolumn{3}{c}{KONIQ-10K / BID} \\
    \cmidrule(r){5-7}
    \multirow{3}{5em}{Authentic} &\checkmark & & 0.26M & 0.716 & 0.673 & 0.495 \\
    &\checkmark & & 2.60M & 0.726 & 0.673 & 0.501 \\
    & &\checkmark & 2.31M & \textbf{0.736} & \textbf{0.680} & \textbf{0.505} \\
    \bottomrule
  \end{tabular}
\end{table}

In this experiment, we employed MobileNetV2 for training and utilized a standard loss function, which is a weighted combination of MSE and PLCC loss, measured between images in the batch. All other settings, such as the learning rate, remained constant. For the within-dataset evaluation, each dataset, KADID-10K and PIPAL, were individually used for both training and testing using 10-fold cross-validation. The performance metrics are reported based on the average of the 10-fold cross-validation results. For the cross-dataset evaluation, we used two train/test pairs: KADID-10K / TID2013 for the Synthetic model and KONIQ-10K / BID for the Authentic model. The results of the within-dataset evaluation on KADID-10K and PIPAL are presented in Table~\ref{tab:single-head}. 
These results indicate that incorporating KAN as the regression head yields improvements over MLP. 
As previously discussed, we used two MLP regression heads: one with the same architecture and another with the same number of FLOPs. In the second experiment, we included the FLOPs for each MLP model and compared them to KAN.
Table~\ref{tab:cross-head} presents the results of the cross-dataset evaluation, showing an improvement when using KAN compared to MLP, highlighting KAN's superior generalizability. Hence, for the rest of the experiments, we only use KAN for quality regression task. It is important to note that making a fair comparison between KAN and MLP is challenging due to the differences in network topology and activation function choices, which significantly impact MLP performance. Similarly, for KAN, factors such as the order of the spline and the number of spline intervals affect the results. Nevertheless, this paper demonstrates that incorporating KAN as part of the regression head can be as effective as using an MLP head.

    \begin{table}[tb]
        \caption{Comparing the performance of different image encoders. Among the evaluated encoders, MobileNetV3 consistently outperforms the others across all metrics.}
        \label{tab:image-enc}
        \centering
        \setlength{\tabcolsep}{6pt}
        \begin{tabular}{@{}lcccccc@{}}
            \toprule
            Model &Backbone & \#Parameters & PLCC & SRCC & KRCC \\
            \midrule
            & &\multicolumn{4}{c}{KADID-10K / TID2013} \\
            \cmidrule(r){3-7}
            \multirow{4}{5em}{Synthetic} & MobileViT-S & 5.6M & 0.61 & 0.59 & 0.42 \\
            & MobileCLIP-S2~\cite{vasu2024mobileclip} & 35.5M & 0.69 & 0.65 & 0.46 \\
            & MobileNetV2~\cite{sandler2018mobilenetv2} & 3.5M & 0.71 & 0.66 & 0.50 \\
            & MobileNetV3~\cite{howard2019searching} & 7.1M &\textbf{0.72} &\textbf{0.68} &\textbf{0.50} \\
            \midrule
            & &\multicolumn{4}{c}{KONIQ-10K / BID} \\
            \cmidrule(r){3-7}
            \multirow{4}{5em}{Authentic} & MobileViT-S & 5.6M & 0.71 & 0.65 & 0.47 \\
            & MobileCLIP-S2 & 35.5M & 0.78 & 0.75 & 0.57 \\
            & MobileNetV2 & 3.5M & 0.74 & 0.68 & 0.51 \\
            & MobileNetV3 & 7.1M &\textbf{0.79} &\textbf{0.78} &\textbf{0.60} \\
            \bottomrule
        \end{tabular}
    \end{table}

\subsection{Comparing Image Encoders} \label{sec:backbones}

Using KAN as the regression head, we now compare different image encoders as the backbone network for our model. Similar to previous experiments, the Synthetic model is trained on KADID-10K and tested on TID2013, while the Authentic model is trained on KONIQ-10K and tested on BID. The performance metrics, including, PLCC, SRCC, and KRCC, are detailed in Table~\ref{tab:image-enc}. The results show that both CNN-based models outperform the Mobile ViT models. The decreased performance of MobileCLIP-S2 can be attributed to the limited training data, as ViT-based models typically require large datasets to effectively learn the extensive network parameters. However, as reported in the next section, we observe that MobileCLIP-S2 helps in better generalization when using multiple color spaces. As expected, MobileNetV3 performs better than MobileNetV2; therefore, MobileNetV3 is used for the rest of our model design.

\subsection{Comparing Various Loss Functions} \label{sec:loss}

\begin{table}[tb]
    \centering
    \caption{Performance comparison of various loss functions considering different color spaces. Best performing model is shown in Bold.}
    \label{tab:loss_performance}
    \resizebox{\textwidth}{!}{%
    \begin{tabular}{@{}lcccccccccccc@{}}
        \toprule
        \textbf{Model} & \multicolumn{2}{c}{\textbf{Loss}} & \multirow{2}{5em}{\textbf{Backbone}} & \multicolumn{3}{c}{\textbf{RGB}} & \multicolumn{3}{c}{\textbf{YUV}} & \multicolumn{3}{c}{\textbf{LAB}} \\
        \cmidrule(r){2-3} \cmidrule(r){5-7} \cmidrule(r){8-10} \cmidrule(r){11-13}
        & \textbf{$\mathcal{L}_{acc}$} & \textbf{$\mathcal{L}_{rob}$} & & \textbf{PLCC} & \textbf{SRCC} & \textbf{KRCC} & \textbf{PLCC} & \textbf{SRCC} & \textbf{KRCC} & \textbf{PLCC} & \textbf{SRCC} & \textbf{KRCC} \\
        \midrule
        & & & & \multicolumn{9}{c}{\textbf{KADID-10K / TID2013}} \\
        \cmidrule(r){5-13}
        \multirow{2}{5em}{\textbf{Synthetic}} & \checkmark & & \multirow{2}{6em}{\centering \textbf{Mobile-\\NetV3}} & 0.72 & 0.68 & 0.50 & 0.41 & 0.34 & 0.25 & 0.47 & 0.39 & 0.28 \\
        & \checkmark & \checkmark & & \textbf{0.73} & \textbf{0.70} & \textbf{0.53} & 0.49 & 0.47 & 0.35 & 0.55 & 0.52 & 0.37 \\
        \cdashline{2-13}
        \multirow{2}{5em}{\textbf{Synthetic}} & \checkmark & & \multirow{2}{6em}{\centering \textbf{Mobile\\CLIP-S2}} & 0.69 & 0.65 & 0.46 & 0.44 & 0.38 & 0.29 & 0.46 & 0.40 & 0.29 \\
        & \checkmark & \checkmark & & 0.69 & 0.66 & 0.48 & \textbf{0.58} & \textbf{0.54} & \textbf{0.39} & \textbf{0.59} & \textbf{0.55} & \textbf{0.40} \\
        
        \midrule
        & & & & \multicolumn{9}{c}{\textbf{KONIQ-10K / BID}} \\
        \cmidrule(r){5-13}
        \multirow{2}{5em}{\textbf{Authentic}} & \checkmark & & \multirow{2}{6em}{\centering \textbf{Mobile-\\NetV3}} & 0.79 & 0.78 & 0.60 & 0.59 & 0.58 & 0.42 & 0.59 & 0.57 & 0.42 \\
        & \checkmark & \checkmark & & \textbf{0.82} & \textbf{0.79} & \textbf{0.61} & 0.69 & 0.68 & 0.51 & \textbf{0.76} & \textbf{0.74} & \textbf{0.56} \\
        \cdashline{2-13}
        \multirow{2}{5em}{\textbf{Authentic}} & \checkmark & & \multirow{2}{6em}{\centering \textbf{Mobile-\\CLIP-S2}} & 0.78 & 0.74 & 0.56 & 0.65 & 0.62 & 0.46 & 0.67 & 0.62 & 0.46 \\
        & \checkmark & \checkmark & & 0.78 & 0.77 & 0.58 & \textbf{0.74} & \textbf{0.72} & \textbf{0.53} & 0.74 & 0.73 & 0.54 \\
        \bottomrule
    \end{tabular}%
    }
\end{table}

\begin{figure}[ht!] 
  \centering 
  \vspace{0.01in}
  \includegraphics[width=\linewidth]{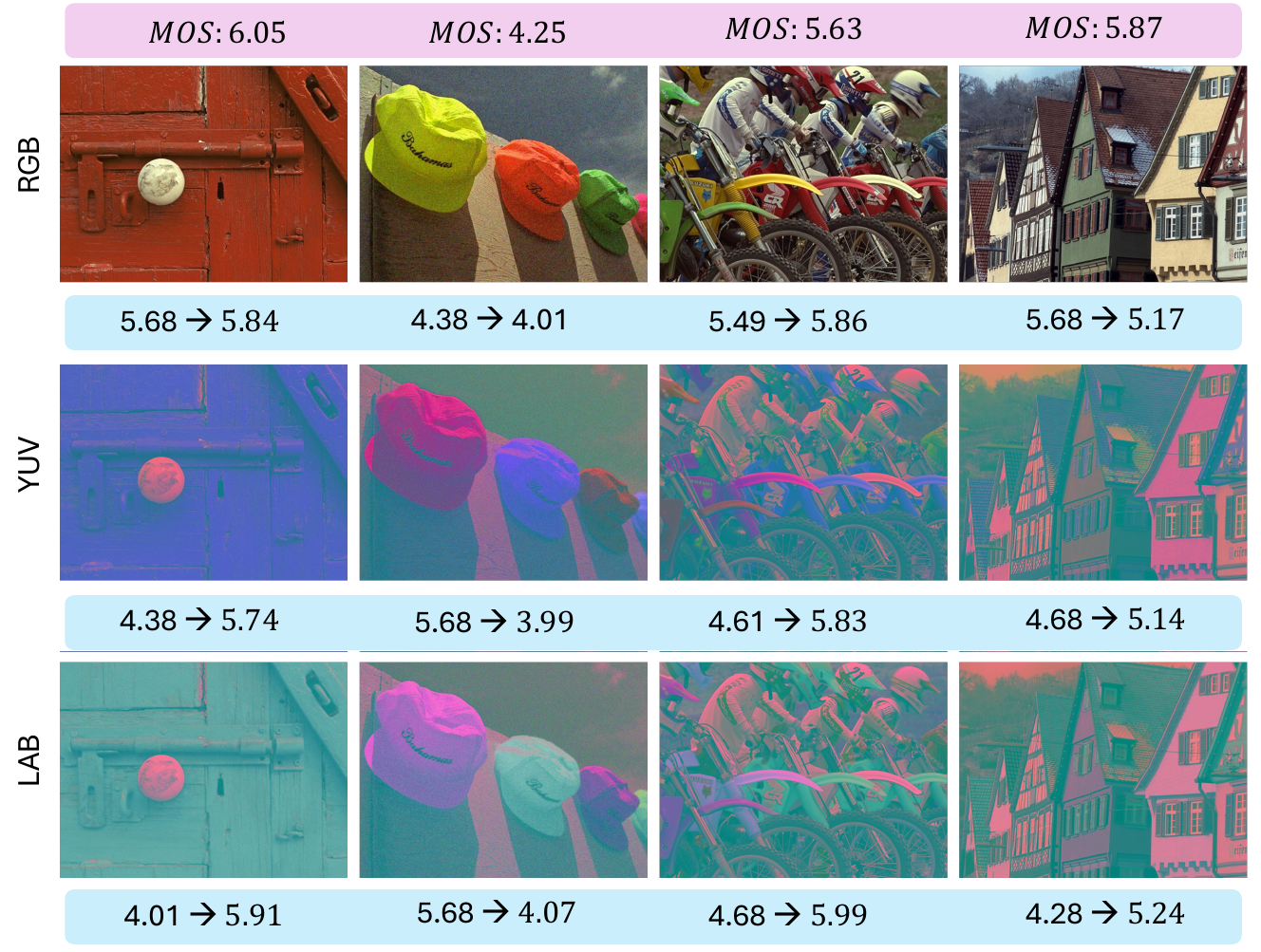} 
\caption{MOS predictions of the Synthetic model using MobileNetV3, trained on KADID-10K and tested on sample images from the TID2013 dataset across different color spaces. Actual MOS scores are highlighted above in the pink colored bar. The change in predicted MOS scores before and after pre-training with \textit{color space loss} ($\mathcal{L}_{rob}$) across various color spaces is shown in the blue colored bar. One can note that the predicted MOS scores for various images across different color spaces converge, indicating the robustness of the metric to different color spaces.}
  \label{fig:color} 
\end{figure}

Table~\ref{tab:loss_performance} presents the performance of various loss functions considering different color spaces (RGB, YUV and LAB). The results indicate that incorporating both \textit{MSE+PLCC} loss ($\mathcal{L}_{acc}$) and the \textit{color space loss} ($\mathcal{L}_{rob}$) improves the model's performance. While the improvement in the RGB color space was not significant, major improvements were observed in the performance metrics for the YUV and LAB color spaces. The MobileCLIP-S2~\cite{vasu2024mobileclip} model, which uses the ViT architecture, is more robust and shows better performance across different color spaces. This could be attributed to the nature of ViT models, which normalize pixel values multiple times during processing.

Figure~\ref{fig:color} shows the MOS predictions of the Synthetic model with MobileNetV3, trained on KADID-10K and tested on sample images from the TID2013 dataset. Despite the differences between the predicted MOS and ground truth values caused by varying scales and types of distortion in the two datasets, training with color space loss ($\mathcal{L}_{rob}$) enhances robusntess and helps the model generalizes better by ensuring that the results remain consistent across different color spaces.

\subsection{Integrating Synthetic and Authentic Branches} \label{sec:syn_auth}

\begin{table}[t]
      \setlength{\tabcolsep}{2pt} 
      \caption{Performance comparison of different training settings for LAR-IQA model on the validation set of the UHD-IQA dataset.}
      \label{tab:syn_auth}
      \centering
      \begin{tabular}{@{}ccccccc@{}}
        \toprule
        \multicolumn{2}{c}{Branch} &\multirow{2}{4em}{Finetune} & \multirow{2}{4em}{KAN neurons} &\multirow{2}{4em}{PLCC} & \multirow{2}{4em}{SRCC} & \multirow{2}{4em}{KRCC} \\
        \cmidrule(r){1-2}
        Synthetic &Authentic & & & & & \\
        \midrule
        \checkmark & & &128 & 0.492& 0.463&0.333 \\
         &\checkmark & &128 & 0.512 & 0.496 & 0.348   \\
        \checkmark &\checkmark & &128 & 0.726 & 0.722 & 0.532 \\
        \hdashline
        \checkmark &\checkmark & &512 & 0.744 & 0.734 & 0.545 \\
        \checkmark &\checkmark &\checkmark &512 & \textbf{0.809} & \textbf{0.803} & \textbf{0.611} \\
        \bottomrule
      \end{tabular}
    \end{table}

In the previous section, we outlined the training process of the Synthetic branch using the KADID-10K dataset and the Authentic branch with the KONIQ-10K dataset. However, these datasets are limited in terms of number of images and distortions types. To enhance our model's performance, we extended our training to include multiple datasets for each branch following a multi-task training process. Specifically, the Synthetic branch was trained on the KADID-10K, TID2013, and PIPAL datasets, while the Authentic branch was trained on the KONIQ-10K, BID, and SPAQ datasets. As outlined in Section 2.2, we mitigated subjective biases by integrating two branches trained on different distortion types, each with a distinct regression head for the respective datasets. We held out a random 10\% of each dataset during the training process for the validation set and model selection per branch. We refer the reader to our previous paper \cite{MSLIQA}, which details the multi-task training process for the IQA task. 

The pretrained models are merged after removing the regression heads and adding new KAN heads. These new heads first downsample the embeddings of each branch, then concatenate the two branches, and finally add a regression head, as illustrated in Figure~\ref{fig:mainFigure}.

\textbf{Training on UHD-IQA Dataset:} We trained our two-branch model on UHD-IQA training dataset in three steps. First, we froze the two pre-trained branches (authentic and synthetic) and trained the KAN heads that reduced the output from 1000 to 256 (128 per branch) and then to a single output neuron. Second, we employed a larger KAN head with a 1000-dimensional embedding mapped to 512 per branch (1024 in total) and then to one neuron for the regression task. Finally, we fine-tuned the model using the pre-trained weights of the authentic and synthetic branches along with the larger KAN head on the new dataset.

Table~\ref{tab:syn_auth} presents the results of various training strategies on the UHD-IQA dataset~\cite{hosu2024uhd} used in the ECCV AIM challenge\footnote{\url{https://codalab.lisn.upsaclay.fr/competitions/19335}}. The model is trained using the training set and tested on the validation set. As observed, the model with the larger regression head, consisting of 512 layers, delivers improved performance. Furthermore, fine-tuning the model significantly improves the results, achieving a PLCC of approximately $0.81$. Overall, the outcomes are promising. It is important to note that for the Authentic branch, we resized the images to 224x224x3, whereas for the Synthetic branch, we used the full-size image.

We present next the comparative evaluation of our proposed model compared to the baselines as evaluated on the validation and test set. The results for the baseline models are obtained from the original publication in~\cite{hosu2024uhd}. 

\subsection{Comparisons with the State-of-the-art} \label{sec:sota_comparison}

\begin{table}[t]
\centering
\caption{Evaluation of the performance of the baselines on the validation set of UHD-IQA. $\uparrow$ means that higher values are better, $\downarrow$ means that lower values are better. Best and second-best scores are highlighted in bold and underlined, respectively.}
\label{tab:val_evaluation}
\begin{tabular}{lccccccc}
\toprule
\textbf{Method} & \textbf{PLCC$\uparrow$} & \textbf{SRCC$\uparrow$} & \textbf{KRCC$\uparrow$} & \textbf{RMSE$\downarrow$} & \textbf{MAE$\downarrow$} & \textbf{\#Para$\downarrow$} & \textbf{MACs$\downarrow$} \\
\midrule
HyperIQA~\cite{su2020blindly} & 0.182 & 0.524 & 0.359 & 0.087 & 0.055 & 27.3M & \underline{211G} \\
Effnet-2C-MLSP~\cite{wiedemann2023konx} & 0.627 & 0.615 & 0.445 & 0.060 & 0.050 & - & 345G \\
CONTRIQUE~\cite{madhusudana2022image} & 0.712 & 0.716 & 0.521 & \underline{0.049} & \underline{0.038} & 27.9M & 855G \\
ARNIQA~\cite{agnolucci2024arniqa} & 0.717 & 0.718 & 0.523 & 0.050 & 0.039 & 27.9M & 855G \\
CLIP-IQA+~\cite{wang2023exploring} & 0.732 & 0.743 & 0.546 & 0.108 & 0.087 & 102M & 895G \\
QualiCLIP~\cite{agnolucci2024quality} & \underline{0.752} & \underline{0.757} & \underline{0.557} & 0.079 & 0.064 & 102M & 901G \\
\midrule
LAR-IQA (MLP head) & \underline{0.797} & \underline{0.791} & \underline{0.601} &\underline{0.042}  &\underline{0.033}  & 21.2M & \textbf{$\leq$37G} \\
LAR-IQA (KAN head) & \textbf{0.809} & \textbf{0.803} & \textbf{0.611} &\textbf{0.040}  &\textbf{0.031}  & 21.1M & \textbf{$\leq$37G} \\
\bottomrule
\end{tabular}
\end{table}

\begin{table}[t]
\centering
\caption{Evaluation of the performance of the baselines on the test set of UHD-IQA. $\uparrow$ means that higher values are better, $\downarrow$ means that lower values are better. Best and second-best scores are highlighted in bold and underlined, respectively.}
\label{tab:test_evaluation}
\begin{tabular}{lcccccc}
\toprule
\textbf{Method} & \textbf{PLCC$\uparrow$} & \textbf{SRCC$\uparrow$} & \textbf{KRCC$\uparrow$} & \textbf{RMSE$\downarrow$} & \textbf{MAE$\downarrow$} & \textbf{MACs(G)$\downarrow$} \\
\midrule
HyperIQA~\cite{su2020blindly} & 0.103 & 0.553 & 0.389 & 0.118 & 0.070 & \underline{211} \\
Effnet-2C-MLSP~\cite{wiedemann2023konx} & 0.641 & 0.675 & 0.491 & 0.074 & 0.059 & 345 \\
CONTRIQUE~\cite{madhusudana2022image} & 0.678 & 0.732 & 0.532 & 0.073 & 0.052 & 855 \\
ARNIQA~\cite{agnolucci2024arniqa} & 0.694 & 0.739 & 0.544 & 0.074 & 0.052 & 855 \\
CLIP-IQA+~\cite{wang2023exploring} & 0.709 & 0.747 & 0.551 & 0.111 & 0.089 & 895 \\
QualiCLIP~\cite{agnolucci2024quality} & 0.725 & 0.770 & 0.570 & 0.083 & 0.066 & 901 \\
\midrule
LAR-IQA (MLP head) & \underline{0.774} & \underline{0.809} & \underline{0.616} &\textbf{0.058}  &\underline{0.042}  & \textbf{$\leq$37} \\

LAR-IQA (KAN head) & \textbf{0.787} & \textbf{0.836} & \textbf{0.642} &\underline{0.061}  &\textbf{0.041}  & \textbf{$\leq$37} \\
\bottomrule
\end{tabular}
\end{table}

Table~\ref{tab:val_evaluation} and Table~\ref{tab:test_evaluation} presents a comparative evaluation result of our proposed model (LAR-IQA) compared to the SOTA models on the AIM UHD-IQA validation and test set respectively. It can be observed that our model LAR-IQA outperforms existing IQA models on both validation and test dataset in terms of all performance measures as well complexity measures. The proposed model is approx. $\times5.7$ faster than the fastest SOTA model, HyperIQA. Furthermore, both branches together have around 21 million parameters, making it suitable for power and resource constrained mobile devices.

It should be noted that in both tables, the term "MLP head" refers to the MLP head with similar FLOPs to the KAN head. This aims to provide the reader with a performance comparison between KAN and MLP on the UHD-IQA dataset.

Additionally, we refer the reader to the UHD-IQA challenge\cite{aim2024uhdbpqa}, which compares various lightweight models proposed for UHD-IQA tasks. In the challenge rankings, LAR-IQA achieved second place. To ensure a fair comparison, we adhered to the challenge's rules and dataset split.

\section{Conclusion} \label{sec:Conc}
To address the lack of low complexity, high accuracy and robust NR-IQA metric, we proposed in this work a lightweight NR-IQA model that surpases the accuracy of current SOTA models while being more suitable for deployment on mobile devices. Our model proposes a dual-branch architecture that processes both authentic and synthetic distortions individually making it more robust to varied distortions as compared to models trained solely on single distortion types.  
Furthermore, we observe that training separately on multiple datasets helps to mitigate subjective biases inherent in individual datasets, thus improving further the model's overall generalizability.

To ensure our model's robustness across various color spaces, we incorporated RGB, YUV, and LAB into the training process. Furthermore, we explored KANs for the quality regression module instead of the commonly used MLPs. Our empirical results show that our proposed model not only surpasses SOTA NR-IQA models in accuracy but is also of much lower complexity, making it well-suited for real-world applications.
\newpage
\bibliographystyle{splncs04}
\bibliography{main}
\end{document}